%% file: acl_latex.tex
\documentclass[11pt]{article}

\usepackage[preprint]{acl}

\usepackage{times}
\usepackage{latexsym}

\usepackage[T1]{fontenc}

\usepackage[utf8]{inputenc}

\usepackage{microtype}

\usepackage{inconsolata}

\usepackage{graphicx}
\usepackage{arydshln}

\usepackage{textcomp}  
\usepackage{scalerel}
\usepackage{amssymb,mathrsfs,amsmath}
\usepackage{xcolor}
\usepackage{pifont}
\usepackage{colortbl}
\usepackage{booktabs}
\usepackage[table]{xcolor} 
\usepackage{array} 
\usepackage{microtype}
\usepackage{inconsolata}
\usepackage{makecell}
\usepackage{multirow}
\usepackage{graphicx}
\usepackage{amsmath}
\usepackage{booktabs}
\usepackage{tabularx}
\usepackage{enumitem}
\usepackage{amsmath, amssymb, amsthm} 
\usepackage{colortbl}
\definecolor{headergray}{gray}{0.92}
\definecolor{wrongred}{RGB}{253,237,237}
\definecolor{vagueorange}{RGB}{255,244,230}
\definecolor{correctgreen}{RGB}{240,248,240}
\definecolor{err1light}{RGB}{255,192,192}
\definecolor{err2light}{RGB}{153,230,153}
\definecolor{err3light}{RGB}{255,227,204}
\definecolor{err4light}{RGB}{153,230,255}
\definecolor{err5light}{RGB}{227,255,153}
\definecolor{err6light}{RGB}{255,153,227}
\definecolor{err7light}{RGB}{255,227,255}
\definecolor{err8light}{RGB}{153,255,227}
\definecolor{rowTraditional}{gray}{0.95}
\definecolor{rowOurs}{RGB}{230,245,255}
\definecolor{rowMerged}{RGB}{235,255,235}

\definecolor{headergray}{gray}{0.92}
\definecolor{keywordgray}{gray}{0.45}
\theoremstyle{definition}

\definecolor{myblue}{RGB}{220,230,250}
\usepackage[most]{tcolorbox}
\newtcolorbox{promptbox}{
    breakable,
    colback=blue!5,
    colframe=blue!50,
    boxrule=0.5pt,
    arc=3pt,
    width=\columnwidth,   
    left=4pt,             
    right=4pt,            
    top=4pt,              
    bottom=4pt,           
    enhanced              
}

\definecolor{myBlue}{HTML}{1f77b4}
\definecolor{myGreen}{HTML}{2ca02c}
\definecolor{myOrange}{HTML}{ff7f0e}
\definecolor{myRed}{HTML}{d62728}

\usepackage{listings}
\usepackage{xcolor}

\lstdefinestyle{aclprompt}{
    backgroundcolor=\color{gray!10},   
    basicstyle=\ttfamily\small,        
    breaklines=true,                   
    columns=fullflexible,              
    frame=single,                      
    framerule=0.5pt,                   
    xleftmargin=1em, xrightmargin=1em, 
    keywordstyle=\color{blue},         
    commentstyle=\color{green!50!black},
    showstringspaces=false,
    tabsize=2,
}

\title{Judge Like Human Examiners: A Weighted Importance Multi-Point Evaluation Framework for Generative Tasks with Long-form Answers}

\author{Guoxin Yu${}^{\dagger 1}$, Chulun Zhou${}^{\dagger 2}$, Lemao Liu${}^{* 3}$, Qi Wang${}^{1,5}$, Mo Yu${}^{}$,\\ \textbf{Jialong Tang${}^{4}$, Baosong Yang${}^{4}$, Xiang Ao${}^{5}$, Wai Lam${}^{2}$, Yue Yu${}^{* 1}$} \\ \\
${}^{1}$ Pengcheng Laboratory, Shenzhen, China.\\
${}^{2}$The Chinese University of Hong Kong. \\
${}^{3}$Fudan University.\quad ${}^{4}$Qwen team, Alibaba Group.\\
${}^{5}$Institute of Computing Technology, CAS.\\
\texttt{\{yugx, yuy\}@pcl.ac.cn}\\
}

\begin{document}
\maketitle
\def\thefootnote{\arabic{footnote}}
\def\thefootnote{$\dagger$}\footnotetext{Equal contribution.}\def\thefootnote{\arabic{footnote}}
\def\thefootnote{*}\footnotetext{Corresponding authors.}\def\thefootnote{\arabic{footnote}}
\begin{abstract}
Evaluating the quality of model responses remains challenging in generative tasks with long-form answers, as the expected answers usually contain multiple semantically distinct yet complementary factors that should be factorized for fine-grained assessment. 
Recent evaluation methods resort to relying on either task-level rubrics or question-aware checklists. However, they still 1) struggle to assess whether a response is genuinely grounded in provided contexts; 2) fail to capture the heterogeneous importance of different aspects of reference answers. Inspired by human examiners, we propose a Weighted Importance Multi-Point Evaluation (WIMPE) framework, which factorizes each reference answer into weighted context-bound scoring points. Two complementary metrics, namely Weighted Point-wise Alignment (WPA) and Point-wise Conflict Penalty (PCP), are designed to measure the alignment and contradiction between model responses and reference answers. Extensive experiments on 10 generative tasks demonstrate that WIMPE achieves higher correlations with human annotations. 
\end{abstract}
\input{0-intro}
\input{1-eval_method}
\input{2-exp1}
\input{3-exp2}
\input{5-conclusion}

\bibliography{custom,6-evaluation}

\appendix
\input{7-appdix_settings}

\end{document}

%% file: 0-intro.tex
\section{Introduction}\label{sec:intro}
With the rapid advancement of large language models (LLMs), evaluating the quality of LLMs' responses in generative tasks with long-form answers has become a fundamental yet challenging problem~\cite{10.1145/3485766,bai-etal-2024-longbench,an-etal-2024-l,huang-etal-2025-minilongbench}. In such scenarios, the expected answers usually constitute a composite construct jointly determined by multiple semantically distinct yet complementary factors, which necessitates a fine-grained evaluation for an effective and robust assessment of response quality.

\begin{figure*}[t]
    \centering
    \includegraphics[width=\linewidth]{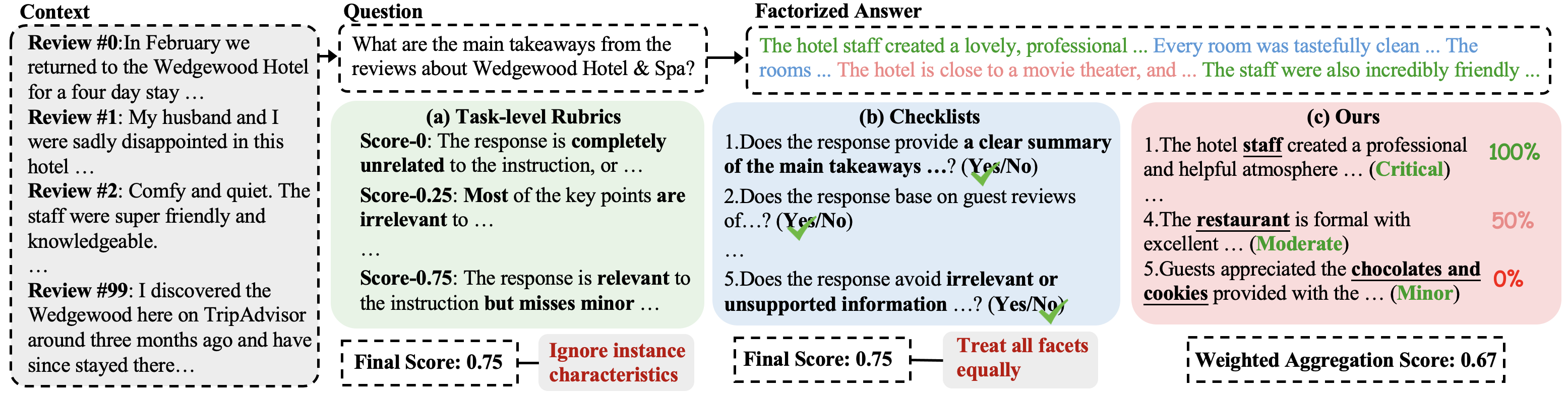}
    \vspace{-5pt}
    \caption{An example of a question with its long-form reference answer given the context, as well as the comparison of evaluation frameworks among \textbf{(a)} task-level rubrics, \textbf{(b)} checklists, and \textbf{(c)} our WIMPE. 
    }
    \label{fig:0-rubrics_comparison}
    \vspace{-10pt}
\end{figure*}

Beyond traditional token-based metrics such as BLEU~\cite{papineni2002bleu} and ROUGE~\cite{lin2004rouge}, recent studies have explored LLM-based automatic metrics to conduct evaluation in generative tasks with long-form answers~\cite{chang2024survey}.
Some works employ task-level rubrics to support pair-wise comparison or direct scoring, which construct general or abstract dimensions (\emph{e.g.} correctness, completeness) to evaluate the response quality across diverse instances~\cite{kim2023prometheus,kim2024prometheus,ye2024flask}. These approaches just carry out simple comparison or coarse scoring without considering instance-specific characteristics. As depicted in Fig.~\ref{fig:0-rubrics_comparison}(a), \citet{kim2024prometheus} enacts the general task-level rubrics composed of multiple score levels.
Moreover, another line of studies introduce a strategy that decomposes the assessing criteria into more concrete question-aware scoring items, taking into account the particular requirements of individual question~\cite{cook2024ticking,que2024hellobench,furuhashi2025checklists,kim2025biggen,wang2025autoscore}. As shown in Fig.~\ref{fig:0-rubrics_comparison}(b), \citet{furuhashi2025checklists} construct checklists comprising multiple question-aware items, each of which requires a binary (\emph{yes}/\emph{no}) judgment.

Although the above evaluation methods have been increasingly applied, the challenge of effectively evaluating response quality remains prominent when the expected answers are long and intrinsically factorized, especially in long-context scenarios~\cite{que2024hellobench,yen2025helmet,zhou-etal-2025-essence,zhuang-etal-2025-self}. 
In fact, existing methods relying on coarse task-level rubrics or question-aware checklists overlook the core requirements for fine-grained assessment and fail to exploit the structured information within long-form answers from the following two perspectives: (1) a high-quality response must directly address the question grounded in the provided context without conflicting content; (2) the heterogeneous importance of different aspects from long-form answers should be distinguished to measure the overall response quality. In this sense, such limitations undermine the reliability of current evaluation approaches, leaving their results prone to assessment biases like criteria entanglement~\cite{li-etal-2025-curse}.

Inspired by human examiners’ grading practices based on predetermined criteria with heterogeneously important scoring points, we propose a \underline{\textbf{W}}eighted \underline{\textbf{I}}mportance \underline{\textbf{M}}ulti-\underline{\textbf{P}}oint \underline{\textbf{E}}valuation~(WIMPE) framework.
As shown in Fig.~\ref{fig:0-rubrics_comparison}(c), our WIMPE resorts to factorizing long-form answers into multiple context-bound scoring points, each of which is assigned one of the three-level weights representing its importance. Based on the factorized scoring points and corresponding importance weights, we design two complementary metrics, namely Weighted Point-wise Alignment~(WPA) and Point-wise Conflict Penalty~(PCP). 
WPA quantifies the weighted alignment between a response and the reference, whereas PCP measures a response's contradictions with the reference.
In this way, WIMPE could more flexibly support more fine-grained assessment and distinguish the heterogeneous importance of different aspects of long-form answers to facilitate more effective and robust evaluation. 

We validate the effectiveness of our WIMPE framework in terms of the scoring correlation with human judgments using 10 existing generative benchmarks with long-form answers. To reduce the burden of manual annotations, we present a \underline{\textbf{ST}}r\underline{\textbf{A}}tified \underline{\textbf{R}}anking method~(STAR) to yield LLM-based relative rankings of inference results of $10$ advanced LLMs. Experiments show that WIMPE achieves higher correlations compared with existing automatic methods. More in-depth analyses exploit the explanations generated during WIMPE evaluation to achieve fine-grained performance comparison and error attribution.
To address the efficiency concerns, we also show that training a lightweight evaluator can effectively support training for WIMPE. 
Our contributions are summarized as follows:
\begin{itemize}[leftmargin=1em,noitemsep]
    \item We proposed an importance weighted multi-point evaluation framework for factorized answers, comprising two complementary metrics, WPA and PCP, that exhibit stronger correlations with response quality validated by LLM-based stratified ranking.
    \item We conducted an interpretability analysis based on the explanations produced by WIMPE for different LLMs in long context comprehension.
    \item We will release the constructed scoring points and their associated importance weights for widely used generative benchmarks, thereby better exploiting existing reference answers by transforming them into structured.
\end{itemize}

%% file: 1-eval_method.tex
\section{Evaluation Approach}\label{Methods}

\begin{figure*}
    \centering
    \includegraphics[width=\linewidth]{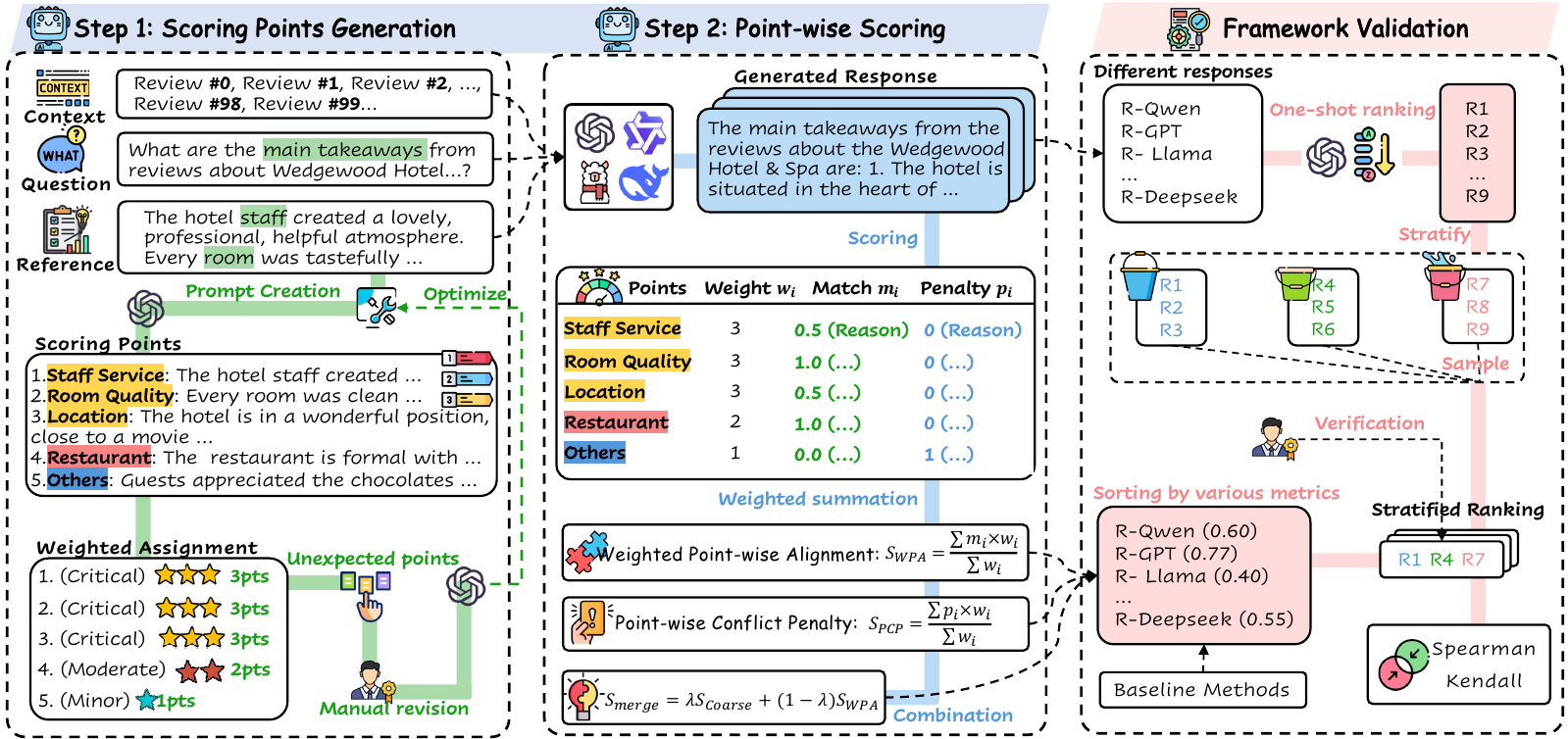}
    \caption{Procedures of Weighted Importance Multi-Point Evaluation (WIMPE) and framework validation. Step 1 generates the weighted scoring points, and Step 2 computes the proposed metrics. Framework validation computes the correlation between evaluation results and LLM-based stratified ranking.}
    \label{fig:wimpe}
    \vspace{-10pt}
\end{figure*}

As illustrated in Fig.~\ref{fig:wimpe}, given a question and its reference answer along with background context, our proposed Weighted Importance Multi-Point Evaluation (WIMPE) framework assesses a model response through the following three steps.
In step~1, WIMPE factorizes the reference answer into multiple weighted scoring points of heterogeneous importance, where each point corresponds to a single aspect of the reference. 
In Step~2, WIMPE compares the response with the factorized scoring points and measures the  Weighted Point-wise Alignment (WPA) and Point-wise Conflict Penalty (PCP), and then computes their combination scores.
In Step~3, A stratified ranking method (STAR) is proposed to generate pseudo-human labels, which are used to analyze and validate the effectiveness of the proposed metrics.


All steps are instantiated using LLMs~\cite{hurst2024gpt} and Appendix~\ref{prompt_engeneering} details the prompts and associated refinement procedures.
Importantly, our primary contribution lies in the evaluation framework and metric design, rather than in prompt construction or optimization.

\subsection{Generation of Factorized Scoring Points}
\label{ScorePointGen}
A \emph{scoring point} encapsulates a single aspect factorized from the reference answer, typically consisting of one or several semantically complete and self-contained sentences. All these scoring points collectively contribute to a comprehensive assessment of response quality with respect to the question. The absence of certain scoring points may compromise the completeness of the answer to varying degrees, depending on their relative importance. 

For a question $q$ and its reference answer $a$, we prompt the LLM to decompose $a$ into fine-grained molecular blocks and summarize them into abstractive \emph{scoring points}:
\begin{equation}\label{eq-1}
    \{s_i\} = \text{LLM}(q, a),
\end{equation}
where $\{s_i\}$ denotes the set of factorized scoring points. The granularity and total number of scoring points are determined by the information density of $a$: higher-density answers yield shorter, more fine-grained scoring points, whereas lower-density answers result in longer, more coarse-grained units.

In this way, the reference answer is transformed into a set of structured and fine-grained scoring points, supporting more precise and interpretable evaluation by explicitly modeling the contribution of individual semantic units.
Furthermore, the proposed formulation is not limited to settings with human-annotated reference answers. Prior work has shown that strong LLMs can generate high-quality proxy references~\cite{oh2023evaluation,ke2024critiquellm}, which can be directly incorporated into our pipeline, extending WIMPE to scenarios where explicit references are unavailable.

\subsection{Three-Level Weight Assignment}\label{ScorePointWeight}
Apart from generating the set of scoring points $\{s_i\}$, an \emph{importance weight} is assigned to each point to represent its importance to the overall answer. Prior studies in the educational measurement literature~\cite{nitko1996educational} have demonstrated that using integer weights within a restricted range (typically $1$--$3$ or $1$--$4$) effectively differentiates the relative significance of assessment components while maintaining structural coherence within a question.
Drawing on this well-established principle, we designed a three-level weighting scheme for determining the importance weight $w_i$ associated with each scoring point $s_i$ as follows:
\begin{equation}
w_i = 
\begin{cases}
3.0, & \text{if } s_i \text{ is Critical}, \\
2.0, & \text{if } s_i \text{ is Moderate}, \\
1.0, & \text{if } s_i \text{ is Minor}.
\end{cases}
\end{equation}
$|\{w_i\}| = |\{s_i\}|$ denotes the number of scoring points, where $s_i$ is the $i$-th point and $w_i$ indicates its importance. Higher weights are assigned to more essential content so that it exerts a correspondingly greater influence on the evaluation outcome.
Such design is also supported by research in educational assessment~\cite{popham1997s,jonsson2007use}, which indicates that a three-level weighting scheme achieves an optimal balance between interpretability and sensitivity.

\subsection{Weighted Point-wise Alignment (WPA)}
\label{WPA}
Building on \S~\ref{ScorePointWeight}, where scoring points are assigned different importance weights, we evaluate the extent to which a generated response covers these points by introducing a \emph{Weighted Point Alignment} (WPA) metric. The core idea is to assess the alignment between a response and each scoring point, and then aggregate these alignment degrees in a weighted manner.
Prior research in educational assessment~\cite{Moskal2000Scoring,jonsson2007use} has demonstrated that rating schemes with too many rating levels tend to blur the boundaries between adjacent categories for raters, whereas rubrics with about three to four levels generally yield the highest scoring reliability. 
Therefore, we adopt a discrete, three-scale alignment rating scheme. For a response $\hat a$, its alignment degree $m_i$ for each scoring point $s_i$ is defined as follows:
\begin{equation}
{m}_i = 
\begin{cases}
 1.0, & \text{if } s_i \text{ is fully covered by } \hat{a}, \\
 0.5, & \text{if } s_i \text{ is partially covered by } \hat{a}, \\
0, & \text{otherwise}.
\end{cases}
\end{equation}
Employing discrete alignment scales enhances scoring robustness by avoiding the arbitrariness and inconsistency associated with selecting continuous numeric values between $[0,1]$.

To enable interpretability for subsequent error analysis, for each scoring point $s_i$,  we prompt LLM to produce not only the alignment degree ${m}_i$ but also a brief explanation $e_i$ for a given response $\hat{a}$:
\begin{equation}\label{eq:match}
    \{m_i, e_i\} = \text{LLM}(q,\{s_i\},\hat{a}).
\end{equation}
The overall alignment score for the instance is then computed as a normalized weighted sum:
\begin{equation}
{S}_{\text{WPA}} = \frac{\sum_{i=1}^{|\{s_i\}|} {m}_i \times w_i}{\sum_{i=1}^{|\{s_i\}|} w_i},
\end{equation}
which ensures comparability across instances with different numbers of scoring points. 



\subsection{Point-wise Conflict Penalty (PCP)}
While the WPA metric captures how well a response covers the required information of the reference, it does not consider to what extent the response introduces conflicting content. To address this, we introduce the \emph{Point-wise Conflict Penalty} (PCP) metric, which explicitly penalizes statements that are irrelevant or contradictory. 
For simplicity, we define the penalty indicator $\hat{p}_i$ using a binary scheme:
\begin{equation}
{p}_i =
\begin{cases}
1.0, & \text{if } \hat{a} \text{ contradicts } s_i, \\
0,   & \text{otherwise}.
\end{cases}
\end{equation}
Similar to WPA, we prompt LLM to generate both the penalty indicator and the corresponding explanations.
The overall penalty is then computed as a normalized weighted sum:
\begin{equation}
S_{PCP} = \frac{\sum_{i=1}^{|\{s_i\}|} \hat{p}_i \cdot w_i}{\sum_{i=1}^{|\{s_i\}|} w_i}.
\end{equation}
Here, ${p}_i$ reflects whether the generated response $\hat{a}$ introduces content conflict with $s_i$. Consequently, contradictions involving more critical scoring points incur greater penalties. This design ensures that the evaluation framework accounts not only for missing information but also for context-violating or misleading content.
\subsection{Combined Metrics}

Since WPA focuses on instance-specific assessment, it may overlook the benefit of task-level rubrics during evaluation. Therefore, we further merge WPA with task-level rubrics to a combined \emph{Merge} score in the following way: 
\begin{equation}\label{eq:merge_score}
    S_{\text{Merge}} = \lambda_m\cdot \, S_{\text{Coarse}} + (1-\lambda_m)\cdot \, S_{\text{WPA}},
\end{equation}
where $S_\text{Coarse}$ denotes the {Coarse 3-level} metric that assesses the reference answer holistically using the task-level evaluation criteria described in \S \ref{exp_setting}.
This merged score fuses both task-level and instance-specific evaluations by a weighted sum, where $\lambda_m \in [0,1]$ controls their respective significance.



\input{exp_results/correlation_on_long_context_QA}

%% file: exp_results/correlation_on_long_context_QA.tex
\begin{table*}[t]
\centering
\small
\setlength{\tabcolsep}{4pt} 
\resizebox{\textwidth}{!}{%
\begin{tabular}{lcccccccccc}
\toprule
\textbf{Metric} 
 & \textbf{StoryQA} & \textbf{ReviewSumm} & \textbf{MeetingSum} &
 \textbf{FinancialQA} & \textbf{PaperAssist}
 & \textbf{ConversMem} & \textbf{ContractQA} & \textbf{LongStory}
 & \textbf{NewsSumm} & \textbf{MultiDocQA} \\
\midrule
\multicolumn{11}{c}{\textbf{Stratified  Instance-level Correlations (Spearman)}} \\
\midrule
\rowcolor{rowTraditional}
BLEU           & 0.5577 & -0.3619 & 0.0714 & -0.1346 & -0.2227 & 0.8500 & 0.2619 & 0.5918 & 0.1818 & 0.0691 \\
\rowcolor{rowTraditional}
ROUGE-L        & 0.6346 & -0.3993 & 0.1261 & -0.1885 & -0.3455 & 0.8000 & 0.2202 & 0.5714 & -0.0455 & 0.0751 \\
Checklist      & 0.6061 & 0.2242 & 0.2953 & 0.1171 & 0.1139 & 0.2943 & 0.4160 & 0.3247 & -0.2756 & 0.2273 \\
Coarse 5-level & 0.6075 & \underline{0.3904} & 0.4315 & 0.2803 & 0.2256 & 0.3110 & \textbf{0.5640} & 0.7636 & 0.1181 & \underline{0.3253} \\
Coarse 3-level & 0.8303 & 0.1607 & 0.3475 & 0.3395 & \underline{0.2256} & 0.7896 & 0.4604 & 0.6636 & 0.0122 & 0.3199 \\
\rowcolor{rowOurs}
WPA (ours)     & \textbf{0.8767} & \textbf{0.4270} & \textbf{0.5823} & \textbf{0.5450} & \textbf{0.4299} & \underline{0.8396} & \underline{0.4696} & \textbf{0.8442} & \underline{0.4545} & \textbf{0.3381} \\
\rowcolor{rowMerged}
PCP (ours)     & \underline{0.8125} & 0.2357 & \underline{0.5043} & \underline{0.4978} & 0.2248 & \textbf{0.8806} & 0.4719 & \underline{0.7986} & \textbf{0.4895} & 0.2576 \\
\midrule
Merge (Ours) & {0.9203}$^{\color{blue}\uparrow}$ & {0.4383}$^{\color{blue}\uparrow}$ & {0.5896}$^{\color{blue}\uparrow}$ & {0.5296} & {0.4687}$^{\color{blue}\uparrow}$ & {0.8729}$^{\color{blue}\uparrow}$ & {0.5268}$^{\color{blue}\uparrow}$ & {0.8316} & 0.4318 & {0.3920}$^{\color{blue}\uparrow}$ \\
\midrule

\multicolumn{11}{c}{\textbf{Stratified  Instance-level Correlations (Kendall)}} \\
\midrule
\rowcolor{rowTraditional}
BLEU          & 0.4872 & -0.3371 & 0.0833 & -0.1179 & 0.2227 & 0.8000 & 0.2381 & 0.5442 & 0.1515 & 0.0510 \\
\rowcolor{rowTraditional}
ROUGE-L       & 0.5897 & 0.3607 & 0.1155 & -0.1718 & -0.3030 & 0.7778 & 0.2103 & 0.5442 & -0.0606 & 0.0591 \\
Checklist     & 0.5877 & 0.2108 & 0.2744 & 0.1093 & 0.1059 & 0.2694 & 0.3900 & 0.2993 & -0.2446 & 0.2132 \\
Coarse 5-level& 0.5647 & \underline{0.3680} & 0.4066 & 0.2674 & 0.2121 & 0.2916 & \textbf{0.5273} & 0.7268 & 0.1265 & \underline{0.3055} \\
Coarse 3-level& \underline{0.7832} & 0.1490 & 0.3308 & 0.3256 & \underline{0.2121} & 0.6436 & 0.4293 & 0.6339 & 0.0167 & 0.3004 \\
\rowcolor{rowOurs}
WPA (ours)    & \textbf{0.8467} & \textbf{0.3813} & \textbf{0.5413} & \textbf{0.5203} & \textbf{0.3941} & \underline{0.7938} & 0.4337 & \textbf{0.8058} & \underline{0.4242} & \textbf{0.3059} \\
\rowcolor{rowMerged}
PCP (ours)    & 0.7602 & 0.2071 & 0.4717 & \underline{0.4779} & 0.1845 & \textbf{0.8371} & \underline{0.4364} & \underline{0.7546} & \textbf{0.4561} & 0.2455 \\
\midrule
Merge (Ours)& {0.8923}$^{\color{blue}\uparrow}$ & {0.3891}$^{\color{blue}\uparrow}$ & {0.5454}$^{\color{blue}\uparrow}$ & {0.5050} & {0.4247}$^{\color{blue}\uparrow}$ & {0.8271}$^{\color{blue}\uparrow}$ & {0.4785}$^{\color{blue}\uparrow}$ & {0.7891} & 0.3647 & {0.3647}$^{\color{blue}\uparrow}$ \\
\bottomrule
\end{tabular}%
} 
\caption{Instance-level correlations with stratified rankings. Across all metrics except \emph{Merge}, \textbf{Best} results are in bold, and \underline{second-best} are underlined. $^{\color{blue}\uparrow}$ signifies that Merge shows improvement to WPA.
}
\label{tab:long-context-corr}
\vspace{-10pt}
\end{table*}

%% file: 2-exp1.tex
\section{Experiment}
\subsection{Settings}\label{exp_setting}
\textbf{Datasets.} To validate the effectiveness of the WIMPE framework, we conduct experiments on ten generative task datasets with long-form answers~\cite{an-etal-2024-l,qiu2024clongeval,tang2024citeeval}. These tasks require models to generate a response $a$ to a given question $q$ based on a given context $c$. Please refer to Appendix~\ref{dataset_details} for more details. 
\noindent\textbf{Baselines.}
We compare {BLEU}~\citep{papineni2002bleu} and {ROUGE-L}~\citep{lin2004rouge} as conventional token-based methods, and select a task-level {Coarse 5-level} criteria and an instance-specific {Checklist} approach from \citet{furuhashi2025checklists}.
Inspired by \citet{kim2024prometheus,kim2025biggen}, we also compare a context-bound {Coarse 3-level} rubric. More details could be found in Appendix~\ref{appdix:baselines}.



\subsection{Experimental settings}
\label{star_method}
To compare among various evaluation methods, we measure the correlations between the evaluation results of different approaches and response quality. As it is difficult to directly describe the absolute quality of a response, we turn to acquire the relative rankings of various responses generated by different LLMs. Then, the problem of describing response quality can be transformed into obtaining the relative rankings of various responses, which are then used to calculate their correlation with different evaluation methods. Specifically, we use ten diverse LLMs (listed in Appendix~\ref{appendix:selected_models}), spanning multiple model families, parameter scales, and context capacities, to generate sufficiently diverse responses to the questions of the involved datasets.
\paragraph{LLM-based Stratified Ranking.} Given the substantial cost and inconsistency incurred by manually ranking multiple model responses~\cite{lo-wu-2014-reliability} and the limited reliability of direct LLM-based annotations~\cite{saito2023verbosity,shi2024judging}, we design a {{ST}}rAtified {{R}}anking method (STAR) to generate pseudo rankings. 
We prompt the most advanced GPT-4o to rank $10$ candidate responses from different LLMs and adopt stratified sampling to select those responses with significant ranking gaps, as shown in Fig.~\ref{fig:wimpe}. For each instance, the sorted responses $\{a_{(1)}, ..., a_{(10)}\}$ could be partitioned into $L$ disjoint groups $\{\mathcal{S}_1, \dots, \mathcal{S}_L\}$ such that each group contains $\lfloor 10 / L \rfloor$ responses. Then we select the fixed $n$-th in each group:
\begin{equation}
    \tilde{a}_l = a_{(k_l + n - 1)}, \quad \text{for group } l = 1, \dots, L,
\end{equation}
where \(k_l\) is the starting index of group \(l\) in the sorted list.
The final pseudo ranking set is $\tilde{\mathcal{A}} = \{\tilde{a}_1, \dots, \tilde{a}_L\}$.
In this way, STAR produces reliable relative rankings that indirectly reflect response quality, thereby mitigating the instability of LLM-based ordering observed for closely matched texts~\cite{shi2024judging}.

We validate the reliability of STAR in Appendix~\ref{Auxiliary_Expe_Ana}, while it does not imply that LLM-based ranking can be directly adopted as a general evaluation method. STAR is a validation mechanism that selectively identifies responses with sufficiently large ranking gaps. Directly using LLMs for response ranking still suffers from unstable problems~\cite{shi2024judging}. 

Inspired by \citet{shimorina2018human,kobayashi2024revisiting}, we report both instance-level Spearman’s $\rho$ and Kendall’s $\tau$ using STAR, as detailed in Appendix~\ref{param_config}.
\subsection{Comparison across Evaluation Methods}
In Table~\ref{tab:long-context-corr}, we report the correlations of our WIMPE and baselines with respect to the stratified rankings. The results validate the two key design principles of WIMPE: (i) instance-specific, context-bound evaluation, and (ii) importance-sensitive assessment of heterogeneous information through weighted scoring points.

Methods whose criteria are not explicitly grounded in the reference context consistently underperform most other evaluators. 
{Coarse 5-level} and {Checklist} achieve only marginal improvements over surface-level metrics such as BLEU and ROUGE. 
{Coarse 5-level} relies on task-level guidelines, which limit its ability to adapt to instance-specific contextual constraints. 
Although {Checklist} generates instance-specific items, these items are derived solely from the question rather than the context or reference answer, preventing it from assessing whether a response is context-bound.

Our methods exhibit higher correlations with the stratified rankings of different LLMs' responses on most datasets, highlighting the necessity of context-bound scoring points. {Coarse 3-level} and {Coarse 5-level} fall short of our WIMPE, as they just inspect each response holistically and lack sensitivity to the heterogeneous importance of different fine-grained aspects. 
This comparison implies that decomposing evaluation into weighted scoring points enables finer-grained discrimination of instance-level correctness.

Regarding our two complementary metrics, WPA generally achieves higher correlations than PCP, suggesting that the judgement of relative rankings in generative tasks with long-form answers are more influenced by the coverage of essential information than by the presence of localized contradictions. 
Nevertheless, the \emph{Merge} score, which augments WPA with task-level rubrics, yields improvements over WPA on some datasets, suggesting that task-level rubrics could possibly provide supplementary information beyond instance-specific scoring points.

\subsection{Ablation Test}
\input{ exp_results/ablation_exp}
To further investigate the effect of the weighted point alignment and three-level importance weight assignment scheme, we conduct an ablation study in Table~\ref{tab:ablation}.
The first block (``Score Scale Reduction'') evaluates robustness under controlled score scales, where 3-level and 5-level scales are uniformly mapped to simplified binary scores across different evaluators. Our WPA shows minimal degradation compared with baselines, demonstrating that scoring-point-based evaluation is inherently more robust to coarse-grained scoring scales.

The second block (``Importance Weight Disturbance'') studies the effect of ignoring or disturbing importance differences among scoring points.
We observe noticeable drops for both WPA and PCP across most scenarios, highlighting the necessity of modeling heterogeneous importance.
\vspace{-5pt}

\subsection{Behavioral Analysis}
\begin{figure*}[t]
    \centering
\includegraphics[width=0.95\linewidth]{ 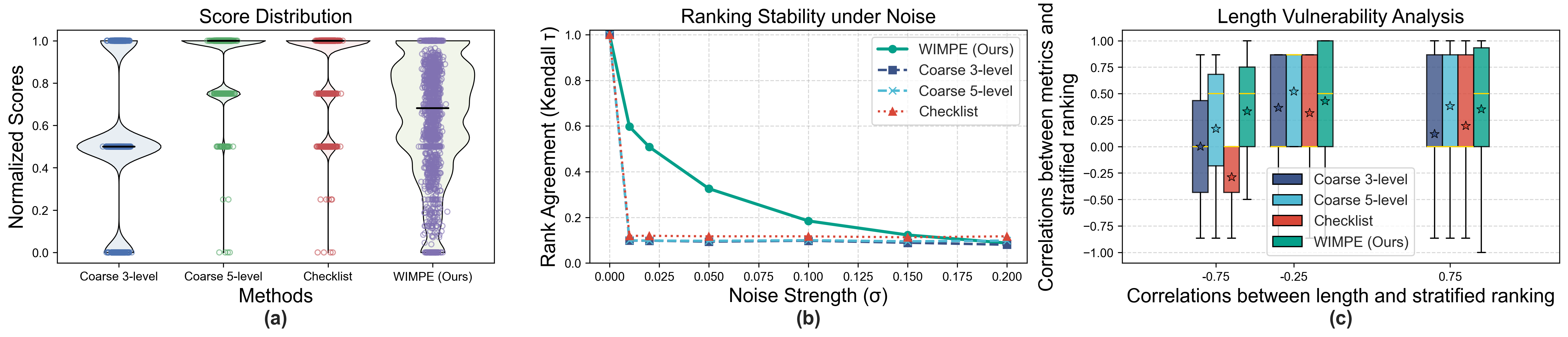}
\vspace{-5pt}
    \caption{Behavioral analysis of different evaluation metrics, including (a) instance-level score distributions, (b) robustness to noise perturbations, and (c) vulnerability to response length. The mean value is explicitly marked with a star symbol in (c). One bin contains no instances and is omitted.}
    \label{fig:behavioral_analysis}
    \vspace{-10pt}
\end{figure*}
To obtain deeper insight into the behavioral characteristics of different evaluation metrics, we conduct a detailed analysis on the Review Summary dataset, which features long-form and factorized answers.

\noindent\textbf{Scores distribution.} As shown in Fig.~\ref{fig:behavioral_analysis} (a), our method exhibits a more finely differentiated score distribution within the same normalized range, whereas baselines are restricted to a small set of discrete values, reflecting their reliance on holistic judgments rather than point-wise aggregation.
Although Checklist and Coarse 5-level nominally support finer scoring scales, their distributions collapse to a few dominant values (mainly 0.5, 0.75, and 1.0). , suggesting that additional score levels are rarely exploited in practice. This observation supports our adoption of a three-level point alignment in \S\ref{WPA}, as increasing scale granularity does not necessarily lead to more effective distinctions.

\noindent\textbf{Robustness to noise.} Fig.~\ref{fig:behavioral_analysis}~(b) shows the correlation with stratified rankings under increasing noise injected into metric scores. As noise intensity increases, our method exhibits a more gradual performance degradation compared to baseline metrics. 
By aggregating reference-grounded scoring points with explicit weights, WIMPE produces rankings driven primarily by genuine quality differences, making them less sensitive to additive perturbations. In contrast, coarse-grained metrics tend to conflate subtly different responses, resulting in more fragile rankings under noise.

\noindent\textbf{Vulnerability to response length.} 
In Fig.~\ref{fig:behavioral_analysis}~(c), we analyze the correlation between WPA and the stratified rankings across bins of response length. In regions where response length exhibits weak correlation with stratified rankings, our method consistently achieves higher median and mean correlations than baselines, indicating its lower vulnerability to superficial length cues. This shows the potential of WIMPE to alleviate length bias.

\section{Interpretability Analysis}
\subsection{Scoring Attribution}
\begin{figure*}[t]
    \centering
    \includegraphics[width=\linewidth]{ 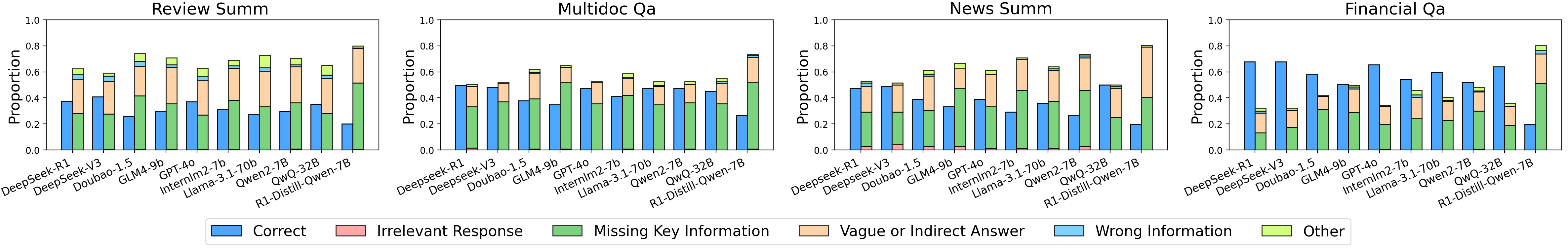}
    \caption{Error type distribution. The right colored bars in each group denote the proportion of each error type.}
    \label{fig:error_distribution}
    \vspace{-10pt}
\end{figure*}
\begin{figure}[t]
    \centering
\includegraphics[width=0.95\linewidth]{ 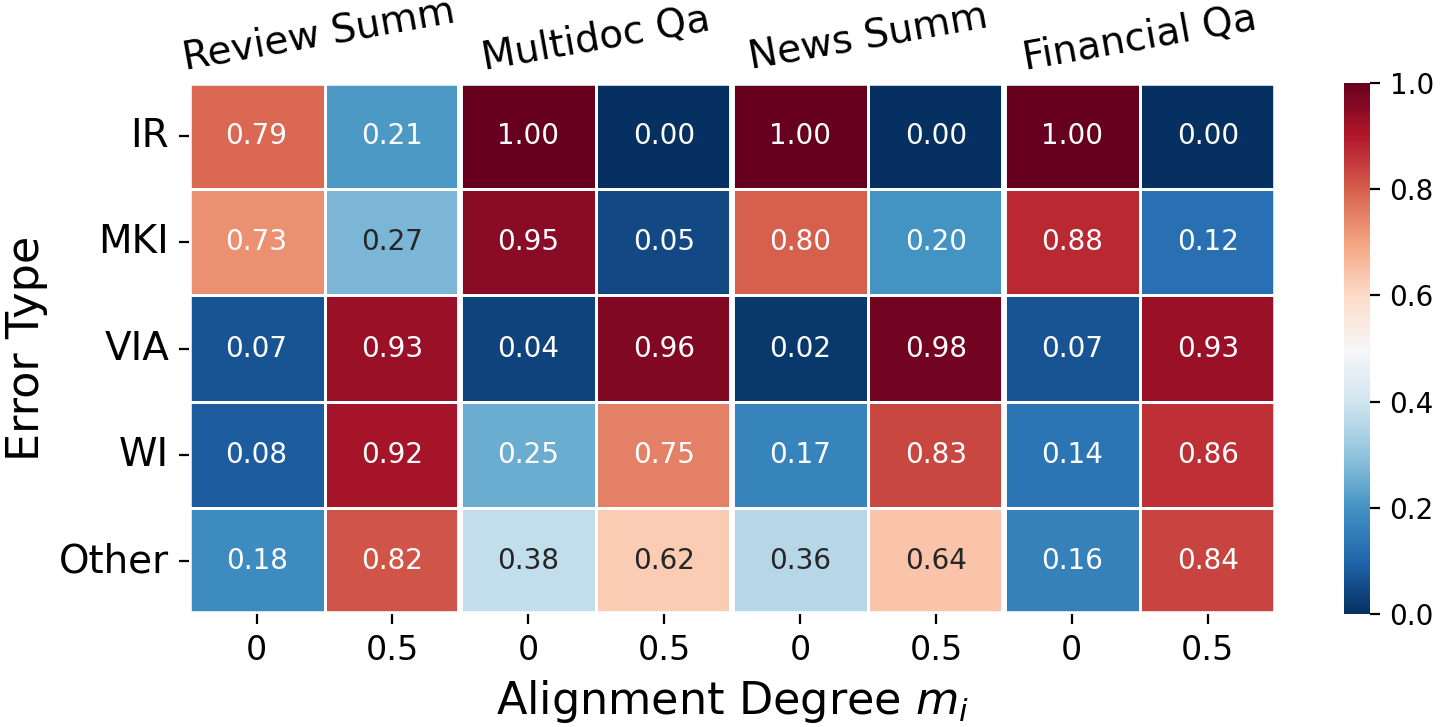}
\vspace{-5pt}
    \caption{Relationship between error types and alignment degrees $m_i$ on different datasets.}
\label{fig:error_score_relationship}
\vspace{-15pt}
\end{figure}
As mentioned in Eq.~\ref{eq:match}, WIMPE generates interpretations $e_i$ in addition to alignment degree $m_i$ when computing WPA. 
Intuitively, the generated explanations can be directly leveraged to attribute alignment scores by revealing why a model response fully, partially, or fails to cover specific scoring points. Based on lexical and semantic cues in these explanations, we categorize non–fully covered cases into the error types defined in Table~\ref{tab:errors_classification}.
We present the distribution of point-wise errors across four datasets and ten evaluated LLMs in Fig.~\ref{fig:error_distribution}. Across all models, the majority of errors are concentrated in \textsc{Missing Key Information} and \textsc{Vague or Indirect Answer}, indicating that even strong LLMs frequently fail to capture all essential elements, resulting in insufficiently context-bound content in generative tasks with long-form answers, especially in long-context scenarios. 
\textsc{irrelevant response} predominantly occurs in the News Summ dataset, likely due to the large number of news items in its context, which makes it easier for LLMs to introduce topic-unrelated information. More analysis could be found in Appendix~\ref{Auxiliary_Expe_Ana}.
Such point-wise attributions remain obscured in existing evaluations. 

For different LLMs, we observe that DeepSeek-R1-distilled-Qwen-7B consistently exhibits the highest error rate, whose error types suggest its poor ability to capture essential information. 
In contrast, DeepSeek-R1/V3 and GPT-4o demonstrate relatively lower error proportions, demonstrating their stronger capabilities for holistic and accurate comprehension.
Notably, although WIMPE is instantiated using GPT-4o, it does not favor GPT-4o-generated responses. We attribute this to the point-wise design, which might mitigate scorer bias and enable model-agnostic assessments that emphasize content-level alignment.

\subsection{Error Severity and Alignment Degree}
To further investigate how WIMPE quantifies error severity, we visualize the distribution of error types across different alignment degrees $m_i$ in Fig.~\ref{fig:error_score_relationship}. 
Across all datasets, we observe a clear pattern: scoring points assigned $m_i=0$ are most associated with \textsc{Missing Key Information} and \textsc{Irrelevant Response}, reflecting severe omissions or off-topic content. 
Conversely, scoring points with $m_i=0.5$ largely correspond to \textsc{Vague or Indirect Answer} and \textsc{Wrong Information}, capturing subtler deficiencies such as incomplete phrasing or minor factual errors.

We can infer that our method considers missing information as severe errors when calculating WPA, and vague or wrong information as moderate errors, which may be regarded as serious errors when calculating PCP.
By aligning the scoring degree with the inherent severity of errors, our method not only quantifies the correctness of responses but also provides interpretable insights into the error types, which is particularly valuable for generative tasks with long-form answers where partial coverage and nuanced inaccuracies frequently occur.
The case study in Appendix~\ref{Auxiliary_Expe_Ana} shows two representative cases and analysis.

%% file: exp_results/ablation_exp.tex
\begin{table}[t]\small
\resizebox{\columnwidth}{!}{
\begin{tabular}{l l c c c}
\toprule
{} & {Method} & FinancialQA & MeetingSum & MultiDocQA \\
\midrule
\multicolumn{5}{c}{Score Scale Reduction (3/5-scale$\rightarrow$ Binary)} \\
\midrule
\multirow{4}{*}{$\rho$} 
 & Coarse-3 & 0.0922{\color{red!70}$\downarrow$} & 0.2100{\color{red!50}$\downarrow$} & 0.1010{\color{red!60}$\downarrow$} \\
 & Coarse-5 & 0.0389{\color{red!70}$\downarrow$} & 0.1602{\color{red!70}$\downarrow$} & 0.1971{\color{red!50}$\downarrow$} \\
 & Checklist & 0.1143{\color{red!20}$\downarrow$} & 0.1830{\color{red!50}$\downarrow$} & 0.1554{\color{red!40}$\downarrow$} \\
 & WPA (ours) & \textbf{0.4821}{\color{red!30}$\downarrow$} & \textbf{0.5539}{\color{red!20}$\downarrow$} & \textbf{0.2491}{\color{red!40}$\downarrow$} \\
\midrule
\multirow{4}{*}{$\tau$} 
 & Coarse-3 & 0.0865{\color{red!90}$\downarrow$} & 0.1980{\color{red!60}$\downarrow$} & 0.0953{\color{red!80}$\downarrow$} \\
 & Coarse-5 & 0.0363{\color{red!90}$\downarrow$} & 0.1512{\color{red!90}$\downarrow$} & 0.1858{\color{red!50}$\downarrow$} \\
 & Checklist & 0.1082{\color{red!10}$\downarrow$} & 0.1725{\color{red!50}$\downarrow$} & 0.1466{\color{red!40}$\downarrow$} \\
 & WPA (ours) & \textbf{0.4671}{\color{red!20}$\downarrow$} & \textbf{0.5110}{\color{red!10}$\downarrow$} & \textbf{0.2369}{\color{red!40}$\downarrow$} \\
\midrule
\multicolumn{5}{c}{Importance Weight Disturbance} \\
\multicolumn{5}{c}{(3-level Weights $\rightarrow$ Average/Random Weights)} \\

\midrule
\multirow{4}{*}{$\rho$}
 & WPA$_{\text{avg}}$ & 0.5398{\color{red!20}$\downarrow$} & 0.5921{\color{red!10}$\uparrow$} & 0.3332{\color{red!30}$\downarrow$} \\
 & PCP$_{\text{avg}}$ & 0.4919{\color{red!20}$\downarrow$} & 0.5017{\color{red!20}$\downarrow$} & 0.2451{\color{red!30}$\downarrow$} \\
 & WPA$_{\text{random}}$ & 0.5255{\color{red!50}$\downarrow$} & 0.5458{\color{red!60}$\downarrow$} & 0.3281{\color{red!30}$\downarrow$} \\
 & PCP$_{\text{random}}$ & 0.5027{\color{red!10}$\uparrow$} & 0.4957{\color{red!20}$\downarrow$} & 0.2307{\color{red!45}$\downarrow$} \\
\midrule
\multirow{4}{*}{$\tau$}
 & WPA$_{\text{avg}}$ & 0.5184{\color{red!20}$\downarrow$} & 0.5566{\color{red!10}$\uparrow$} & 0.3125{\color{red!40}$\downarrow$} \\
 & PCP$_{\text{avg}}$ & 0.4725{\color{red!20}$\downarrow$} & 0.4728{\color{red!10}$\uparrow$} & 0.2327{\color{red!40}$\downarrow$} \\
 & WPA$_{\text{random}}$ & 0.5044{\color{red!50}$\downarrow$} & 0.5038{\color{red!60}$\downarrow$} & 0.3057{\color{red!10}$\downarrow$} \\
 & PCP$_{\text{random}}$ & 0.4793{\color{red!10}$\uparrow$} & 0.4606{\color{red!25}$\downarrow$} & 0.2198{\color{red!50}$\downarrow$} \\
\bottomrule
\end{tabular}}
\caption{Spearman $\rho$ and Kendall $\tau$ results in Ablation Test. $\downarrow$ indicates correlation degradation relative to the full method in Table~\ref{tab:long-context-corr}, $\uparrow$ indicates improvement. Darker arrow colors correspond to larger changes.}
\label{tab:ablation}
\vspace{-10pt}
\end{table}

%% file: 3-exp2.tex
\section{WIMPE with Lightweight Evaluator}

\subsection{Train Light Evaluators}
To obtain reliable WPA scores, our framework necessitates multiple LLM calls to predict point-wise alignment degrees, which can lead to substantial financial costs. 
To address this, we train a lightweight evaluator that replaces GPT-4o in the alignment degree prediction stage, while keeping the subsequent WPA computation unchanged.

We formulate a point-wise alignment prediction task: given a question $q$, a generated response $\hat{a}$, and a scoring point $s_i$, the evaluator predicts an alignment label $y \in \{0, 0.5, 1\}$, indicating the degree of $s_i$ covered in $\hat{a}$, as defined in \S\ref{WPA}. Formally, the evaluator learns a conditional distribution $P_\theta(y \mid q, \hat{a}, s_i)$, where $\theta$ denotes the parameters of the light evaluator. We instantiate this formulation under both encoder-only and decoder-only architectures. For the encoder-only setting, we train a BERT evaluator.
The input sequence is constructed as \texttt{([CLS]$q$[SEP]$\hat{a}$[SEP]$s_i$)} and the final hidden state of \texttt{[CLS]} is projected to a three-dimensional output space.
For the decoder-only setting, we construct an instruction-style input $x = \mathcal{I}(q, \hat{a}, s_i)$ that explicitly describes the scoring semantics, and fine-tune a causal language model to generate the label autoregressively.
The model is optimized using cross-entropy loss over the label tokens. 
\subsection{Experiment and Results Analysis}

\input{ table/light-evaluator}
We collect point-wise alignment degrees data $\{(q,\hat{a},s_i),y\}$ generated by GPT-4o across all datasets, which is split into training, validation, and test sets. The training set is used as supervision for training lightweight evaluators.
We evaluate whether lightweight evaluators can serve as effective substitutes for GPT-4o in the WIMPE pipeline.
Specifically, the correlation comparison measures the impact of replacing GPT-4o on evaluation quality. Accuracy and macro-F1 scores at the alignment degree level quantify how well lightweight evaluators reproduce GPT-4o’s alignment degree. Refer to Appendix~\ref{param_config} for detailed settings.


As shown in Table~\ref{table:light-evaluator}, compared to the BERT-based light-evaluator, the Qwen3-based evaluators achieve accuracy and macro-F1 scores up to $0.6$, indicating a reasonable ability to reproduce GPT-4o’s point-wise alignment judgments.
More importantly, WPA-based rankings computed from lightweight-predicted alignment degrees achieve higher correlations with pseudo-human labels than GPT-4o-based and Coarse 3-level in several cases.
Interestingly, although larger lightweight models obtain higher alignment accuracy, their ranking-level correlations decrease.
This suggests that GPT-generated alignment degrees may contain noise or biases, and overfitting to them could degrade the consistency of WPA-based rankings with human preferences.

Overall, these results demonstrate that cost-efficient lightweight evaluators can effectively replace GPT-4o in predicting alignment degrees within the WIMPE pipeline, substantially reducing evaluation cost while preserving reliable ranking-based assessment.

%% file: table/light-evaluator.tex
\begin{table}[t]\small
\centering
\resizebox{\columnwidth}{!}{
\begin{tabular}{lcccc}
\toprule
Models & \multicolumn{2}{c}{Finetuned Light} & \multicolumn{2}{c}{Vanilla} \\
\cmidrule(lr){2-3} \cmidrule(lr){4-5}
& $\rho$ & $\tau$ & $\rho$ & $\tau$ \\
\midrule
\rowcolor{rowOurs}
WPA$_{\rm GPT\text{-}4o}$ & -- & -- & 0.4708 & 0.4357 \\
\rowcolor{rowTraditional}
Coarse-3$_{\rm GPT\text{-}4o}$ & -- & -- & 0.3139 & 0.2996  \\
\midrule
WPA$_{BERT}$ & 0.1931 & 0.1827 & -- & --   \\
WPA$_{Qw2.5\text{-}0.5B_{Ins}}$ & \textbf{0.4837} & \textbf{0.4462} & -0.0320 & -0.030 \\
WPA$_{Qw3\text{-}1.7B}$ & 0.4591 & 0.4200 & 0.3258 & 0.3107 \\
WPA$_{Qw3\text{-}4B}$ & 0.4801 & 0.4296 & 0.1998 & 0.1916 \\
\midrule
& ACC & F1 & ACC & F1 \\
\midrule
WPA$_{BERT}$ & 0.3402 & 0.5677 & -- & -- \\
WPA$_{Qw2.5\text{-}0.5B_{Ins}}$ & 0.6207 & 0.6161 & 0.3363 & 0.2920 \\
WPA$_{Qw3\text{-}1.7B}$ & {0.6443} & {0.6435} & 0.4444 & 0.4007 \\
WPA$_{Qw3\text{-}4B}$ & \textbf{0.6743} & \textbf{0.6736} & 0.3440 & 0.2804 \\
\bottomrule
\end{tabular}}
\caption{Correlations between different metrics-based and pseudo-human rankings, as well as ACC and macro-F1 scores at the alignment degree level, where GPT-4o-generated alignment degrees are treated as reference. WPA scores are computed using either LLM-predicted (Vanilla) or lightweight-predicted (Finetuned Light) alignment degrees, while all other components remain identical with WIMPE pipeline in Table~\ref{tab:long-context-corr}.}
\label{table:light-evaluator}
\vspace{-15pt}
\end{table}

%% file: 5-conclusion.tex
\section{Conclusion}
Since existing evaluation methods inadequately capture contextual groundedness and heterogeneous importance in generative tasks with long-form answers, we propose a weighted importance multi-point framework with two complementary metrics. Extensive experiments demonstrate the effectiveness of our proposed evaluation framework and better interpretability compared to prior methods. We will release the weighted scoring points to support future research in the community.
\section*{Limitations}
In this paper, we propose a weighted scoring points-based evaluation framework for generative tasks with long-form answers.
Although extensive experiments have proved its effectiveness and interpretability, there are still several limitations.
First, as mentioned in \S~\ref{sec:intro}, our experiments focus on challenging long-context generative tasks. 
Second, our interpretability analysis is limited to comparison among vanilla LLMs, without further error analysis or comparison of specialized long-context methods, such as retrieval-augmented generation.
Third, the lightweight evaluator is trained based on a narrow set of models; although, in theory, any models with considerably smaller parameter scales have the potential to serve as a light scorer, our experiments are only conducted on Qwen and BERT models. We would like to develop a systematic investigation of these limitations in future work.
\section*{Ethical considerations}
All the datasets used in this paper are public and have been widely studied in previous works. We do not identify any ethical concerns with the proposed evaluation method.

%% file: 7-appdix_settings.tex
\input{ 4-related_work}

\input{ 8-Auxiliary_Exp}
\input{ 9-Exp_settings}

%% file: 4-related_work.tex
\section{Related Work}
\subsection{Traditional NLG Evaluation}
Traditional Natural Language Generation (NLG) evaluation methods include n-gram–based metrics, such as BLEU~\cite{papineni2002bleu} and ROUGE~\cite{lin2004rouge}, which assess surface-level overlap between system responses and reference texts. To better capture semantic similarity, embedding-based metrics such as BERTScore~\cite{zhangbertscore}, BARTScore~\cite{yuan2021bartscore}, and MoverScore~\cite{zhao2019moverscore} leverage contextualized representations or model-based likelihoods to evaluate generated text in both reference-based and reference-free settings. More recently, neural trained evaluators have been explored. For example, \citet{zhong2022towards} re-framed NLG evaluation as a dimension-related boolean question answering and introduced an intermediate learning phrase to incorporate external knowledge from multiple related tasks. \citet{xu-etal-2023-instructscore} fine-tuned a LLaMa-based explainable metric by harnessing explicit human instruction and implicit knowledge of GPT-4.
These approaches have been widely used in various text generation evaluations, including long context comprehension~\cite{huang-etal-2025-minilongbench, baek-etal-2025-revisiting}. Nevertheless, they are better suited for tasks with concise and definite answers, which can lead to unfairness or bias in favor of factorized answers.
\subsection{LLM-based NLG Evaluation} 
LLM-based evaluation methods can be broadly categorized according to the granularity of their evaluation criteria. A first line of work adopts task-level evaluation, where a fixed set of rubric dimensions is shared across all instances of a given task. Representative approaches include multi-stage rubric generation and refinement for calibrating LLM evaluators~\cite{liu2024calibrating}, coarse-grained criteria~\cite{zheng2023judging}, skill-based scoring with predefined fine-grained dimensions~\cite{ye2024flask}, rubric-trained evaluators such as Prometheus~\cite{kim2023prometheus,kim2024prometheus,hashemi-etal-2024-llm}, and framework-based scoring approaches like GPTScore~\cite{fu2024gptscore}. An additional study~\cite{lee2024checkeval} decomposed task-level evaluation aspects into sub-components and lightly adapted to each instance.

A second line of work adopts instance-level evaluation, where evaluation criteria are constructed specifically for each instance.
This includes automated checklist construction tailored to individual instructions~\cite{cook2024ticking,linwildbench}, checklists derived from task-level rubrics~\cite {lee2024checkeval}, multiple methods for generating instance-specific rubrics~\cite{furuhashi2025checklists, kim2025biggen, wang2025autoscore}, in which each example is accompanied by its own explicitly defined scoring criteria.

Some of these evaluation methods have been reformed to assess long context comprehension \cite{que2024hellobench,yen2025helmet,zhou-etal-2025-essence,zhuang-etal-2025-self,yu2025preludebenchmarkdesignedrequire,zhou2026improvingmultistepraghypergraphbased} for LLMs. For example, \citet{que2024hellobench} predefined task-level checklists and adopted an LLM to determine whether each checklist is satisfied.
\citet{yen2025helmet} designs different match levels to evaluate the fluency and correctness of generation according to the gold answer.
\citet{zhuang-etal-2025-self} used GPT-4o to judge if the model’s answer is correct from the dimensions of correctness and integrity.
However, most of these methods still regard the evaluated response as a whole and give a holistic judgment, which cannot adapt to the specific requirements of each instance.
Some instance-level criteria, such as checklists, often provide only high-level questions without specifying the concrete information required, and are typically equipped with yes/no binary judgments. 
As a result, answers may be judged correct despite containing flaws that conflicted with the long context or missing crucial information. 
Our method instead derives explicit, weighted scoring points from the reference answer, ensuring that key units are properly emphasized in the final score.

\subsection{Factuality Evaluation}
Factuality Evaluation aims to determine whether a generated response is consistent with established facts, which could effectively detect and mitigate hallucinations in LLMs. Existing studies typically design various self-reflection mechanisms to compare the generated texts with given contexts or retrieved external knowledge~\cite{manakul2023selfcheckgpt,roy2024learning,wang-etal-2025-self-reasoning}.
To achieve more detailed and reliable fact verification, recent works~\cite{min2023factscore,zheng-etal-2025-long-form,yan2025decomposing,yan2025atomic} further decompose generated texts into multiple atomic units and verify each unit separately.

Although such decomposition-based tricks appear similar to ours, their objectives and evaluation targets fundamentally differ.
These methods primarily focus on factual correctness by assessing the alignment between the generated content and supporting evidence.
In contrast, our work centers on fine-grained quality evaluation of factorized answers, where the goal is to assess how well a candidate response meets the criteria implied by its corresponding reference answer across multiple dimensions.

%% file: 8-Auxiliary_Exp.tex
\section{Auxiliary Experiments and Analysis}
\label{Auxiliary_Expe_Ana}
\subsection{Human Validation on the LLM-generated Stratified Rankings} 
To validate the reliability of the acquired relative rankings of responses generated by different LLMs, we recruit two PhD students with domain expertise to manually annotate a randomly sampled subset of the stratified responses, and compensated them at a competitive rate. We find that over 85\% of STAR-induced rankings are consistent with human judgments across different datasets. Most of the discrepancies arise from responses that differ only in minor details or emphasis, which are inherently difficult for humans to order consistently. The results indicate that STAR provides a practical balance between the reliability and cost of LLM-based annotation.
\input{ table/error_types}
\input{ table/cases_for_error_types}

\subsection{Auxiliary Analysis for Error Attribution} 
To save space, we present the error taxonomy in Table~\ref{tab:errors_classification}. 
Accordingly, a higher proportion of \textsc{Other} errors could be observed in the Review Summary dataset in Fig.~\ref{fig:error_distribution}, which contains a large number of questions requiring the summarization of reviews from different individuals about the same hotel.
In these cases, reference summaries typically emphasize the dominant sentiment polarity and omit infrequent or minority opinions (e.g., a few negative comments among largely positive reviews). By contrast, some LLM-generated summaries tend to include such low-frequency evaluations, producing mixed-sentiment summaries that diverge from the expected answers and are consequently labeled as \textsc{Other}.

\subsection{Case Study}
Table~\ref{tab:case_study_combined} shows two illustrative examples of point-wise evaluation and error attribution. Each case presents the evaluation of generated answers against reference answers, covering the question, scoring points, model outputs, assigned scores, and corresponding error types with explanations. The two cases are selected from the Review Summary and News Summary datasets, respectively, which demonstrate how our framework performs fine-grained, point-wise assessment under different content domains.
Taking Question~1 in Table~\ref{tab:case_study_combined} as an example, our method assigns partial credit to responses that capture the general meaning but contain localized factual inaccuracies or incomplete coverage (e.g., incorrect distance or missing mobility cues).
Across both cases, the results demonstrate that our framework can identify fine-grained, scoring-point-level errors and provide faithful attributions, enabling more precise diagnostic analysis beyond holistic evaluation.

%% file: table/error_types.tex
\begin{table*}[t]\small
\centering
\small
\setlength{\tabcolsep}{6pt}
\renewcommand{\arraystretch}{1.3}

\begin{tabular}{p{4cm} p{6.8cm} p{4cm}}
\toprule

\rowcolor{headergray}
\textbf{Error Type} & \textbf{Description} & \textbf{Indicative Keywords} \\
\midrule

\textsc{Irrelevant Response} &
The generated response discusses content that does not address the given question or deviates from the required topic. &
\textcolor{keywordgray}{not answer the question; unrelated topics; which are irrelavant; ...} \\
\midrule
\textsc{Missing Key Information} &
The response omits crucial facts or fails to cover essential elements required by the scoring criteria. &
\textcolor{keywordgray}{does not mention; missing key detail; fails to include; ...} \\
\midrule
\textsc{Vague or Indirect Answer} &
The response only implies the correct information without stating it explicitly, resulting in unclear or incomplete coverage. &
\textcolor{keywordgray}{indirectly imply; not explicit; vague reference; omission; ...} \\
\midrule

\textsc{Wrong Information} &
The response contains statements that directly contradict or incorrectly describe facts specified in the reference. &
\textcolor{keywordgray}{state an incorrect; inconsistency; rather than; unsupported claim; fabricated detail; ...} \\
\bottomrule
\end{tabular}

\caption{Error taxonomy used in our evaluation framework. Each error type is defined with a concise description and indicative lexical cues used for analysis and explanation.}
\label{tab:errors_classification}
\end{table*}

%% file: table/cases_for_error_types.tex
\begin{table*}[htbp]
\centering
\small
\setlength{\tabcolsep}{4pt}
\renewcommand{\arraystretch}{1.3}

\begin{tabularx}{\textwidth}{
    p{3.4cm}
    c
    p{3.9cm}
    c
    p{2cm}
    X
}
\toprule

\multicolumn{6}{l}{\textbf{Question 1:} What are guest opinions on the location of Marquesa Hotel?} \\
\midrule

\rowcolor{headergray}
\textbf{Scoring Point (Excerpt)} &
\textbf{Wt.} &
\textbf{Generated Answer (Excerpt)} &
\textbf{Match} &
\textbf{Error Type} &
\textbf{Explanation (Excerpt)} \\
\midrule

\rowcolor{err4light}
``...a peaceful oasis \textbf{one block} from the bustle of Duval St....''
& 3
& ``...peaceful and quiet seclusion... \textbf{two blocks} away from Duval Street...''
& \textbf{0.5}
& \textsc{Wrong Info}
& ... correctly captures the peaceful atmosphere, \textbf{but states an incorrect distance} to Duval Street. The discrepancy between ``one block'' and ``two blocks'' leads to partial credit. \\

\rowcolor{err3light}
``... within \textbf{walking or biking distance}, even the Old Town attractions ...''
& 3
& ``... \textbf{easy to walk} to local amenities ... central position ...''
& \textbf{0.5}
& \textsc{Vague Indirect Answer}
& While walkability is implied, the generated answer \textbf{omits any reference to biking and does not explicitly mention Old Town attractions}, resulting in incomplete coverage. \\

\rowcolor{white}
``... very convenient location ... walking distance of the Old Town attractions and ... Duval Street ...''
& 2
& ``... central position ... easy access to the hustle and bustle ...''
& \textbf{1.0}
& \textsc{NA}
& The generated answer \textbf{directly aligns} with the reference by explicitly emphasizing convenience and proximity to key attractions. \\

\midrule
\multicolumn{6}{l}{\textbf{Question 2:} Please summarize these news articles.} \\
\midrule

\rowcolor{headergray}
\textbf{Scoring Point (Excerpt)} &
\textbf{Wt.} &
\textbf{Generated Answer (Excerpt)} &
\textbf{Match} &
\textbf{Error Type} &
\textbf{Explanation (Excerpt)} \\
\midrule

\rowcolor{err2light!50}
``... \textbf{There is a trade-off with shovels}: those that make it easier to gather snow make it harder to lift it, and vice versa ...''
& 2
& ``... ergonomic designs and mechanical advantages of specific grips ...''
& \textbf{0}
& \textsc{Missing Key Info}
& ... \textbf{does not mention the trade-off} between shovels that make it easier to gather snow versus lift it ... \\

\rowcolor{err1light}
``... Recommendations to \textbf{reduce risk} include not shoveling while snow is still falling, not bundling up too much, and avoiding coffee or hot cocoa before ...''
& 2
& ``... article emphasizes ergonomic shovel designs ... does not mention safety recommendations ...''
& \textbf{0}
& \textsc{Irrelevant Response}
& ... does {not include safety recommendations}. It \textbf{focuses entirely on shovel grips and ergonomics}, which are \textbf{irrelevant} to this scoring point. \\

\bottomrule
\end{tabularx}

\caption{Case studies illustrating how our evaluation method assigns point-wise scores, explanations, and error categories across two different questions. Row shading highlights different error types, while ellipses indicate omitted content irrelevant to the evaluated scoring point. Long contexts are omitted for concise observation.}
\label{tab:case_study_combined}
\end{table*}

%% file: 9-Exp_settings.tex
\section{Experimental Settings}
\subsection{Datasets Details}\label{dataset_details}
\input{ table/dataset_details}
As described in \S \ref{exp_setting}, we construct an evaluation suite based on open-ended tasks drawn from \citet{an-etal-2024-l}, \citet{qiu2024clongeval}, and \citet{tang2024citeeval}. Specifically, we selected instances with long-form and factorized answers from the original datasets. 
Since WIMPE is designed for instances with factorized and information-rich reference answers, we select those instances that satisfy this criterion. The resulting benchmark spans $10$ datasets, covering various domains and text-generation tasks.
Consequently, the benchmark spans 10 datasets, covering 7 domains: Finance, Law, Meeting Transcripts, Hotel Reviews, Academia, Literature, and Politics. Three types of text-generation tasks are covered: Summarization, Question Answering, and Multi-turn Conversation.
Table~\ref{tab:dataset_stats} reports the detailed number of instances and the covered domain of each dataset.
\subsection{Baselines}\label{appdix:baselines}
We compare WIMPE against several standard and recent evaluation methods.
\textbf{BLEU}~\citep{papineni2002bleu} and \textbf{ROUGE-L}~\citep{lin2004rouge} are included as conventional n-gram overlap metrics.
Among recent LLM-based metrics, we select two representative methods from \citet{furuhashi2025checklists}. \textbf{Coarse 5-level} adopts a five-level scoring scheme by decomposing general criteria into fine-grained rules corresponding to each score level according to task characteristics. 
\textbf{Checklist} also follows a five-level scoring design, and generates question-specific items without consideration of long contexts, evaluating responses according to binary (\emph{yes}/\emph{no}) judgments. 
Since these metrics do not consider the context groundness and thus cannot evaluate whether the generated content adheres to the constraints of the given long context, we create a coarse three-level rubric, \textbf{Coarse 3-level}, inspired by \citet{kim2024prometheus,kim2025biggen}, which assigns each response to one of three quality levels based on its coverage of the reference. 

For three LLM-based baselines, the prompts for Coarse 5-level and Checklist are directly adopted from \citet{furuhashi2025checklists}. Detailed prompt formulations are provided in the original paper, where prompts for Checklist are referred to as ``prompt used in the Specify'', ``Prompt used for direct scoring evaluation with checklists'' and the prompt for Coarse 5-level is denoted as ``Prompt used for direct scoring evaluation without checklists''. 
Inspired by \citet{kim2024prometheus, kim2025biggen}, we created the prompt for ``Coarse 3-level'' which is detailed in Listing.~\ref{lst:prompt_for_coarse_3-level}.
\subsection{Selected Models}\label{appendix:selected_models}
\begin{table}[t]
\centering
\resizebox{\columnwidth}{!}{
\begin{tabular}{lcl}
\toprule
\textbf{Model} & \textbf{\# P} \\
\midrule
InternLM2‑5‑7B‑Chat~\cite{cai2024internlm2} & 7B  \\
Qwen2‑7B‑Instruct~\cite{team2024qwen2} & 7B  \\
DeepSeek‑R1~\cite{guo2025deepseek} & 671B  \\
DeepSeek‑R1‑Distill‑Qwen‑7B~\cite{guo2025deepseek} & 7B  \\
GLM‑4‑9B‑Chat~\cite{glm2024chatglm} & 9B  \\
Doubao‑1.5‑Pro‑32K & NA  \\
DeepSeek‑V3~\cite{liu2024deepseek} & 671B  \\
Qwen‑QwQ‑32B~\cite{yang2025qwen3} & 32B  \\
LLaMA‑3.1‑70B‑Instruct~\cite{grattafiori2024llama} & 70B  \\
GPT‑4o~\cite{hurst2024gpt} & NA  \\
\bottomrule
\end{tabular}}
\caption{Details of the LLMs used for WIMPE evaluation. \#P denotes the parameter scales.}
\label{tab:llm-list}
\end{table}
To assess the correlation between various metrics with pseudo-human labels, we generated ten predictions per instance using diverse LLMs listed in Table~\ref{tab:llm-list}, which support long context inputs. This selection spans multiple model families, parameter scales, and context capacities, providing sufficiently diverse outputs to yield robust experimental results.
\subsection{Parameters Configuration}\label{param_config}
\paragraph{Stratified rankings generation.}
For each instance, we set the number of groups to $L=3$ in the proposed STAR method. Within each layer, we apply stratified sampling to construct stratified rankings by selecting responses at fixed positions, resulting in two sets of ranked responses corresponding to original ranking indices [0,4,8] and [1,5,9].
We compute instance-level correlations of various evaluation methods and the obtained stratified rankings, where the correlation for each instance is calculated separately and then averaged across all instances, following prior work~\cite{louis2013automatically,shimorina2018human,kobayashi2024revisiting}. System-level correlation is not applicable in this setting, as the compared candidate sets vary across instances under the STAR setting described in Section~\S\ref{star_method}.
Across all datasets, we evaluate $702$ instances in total. Since two set of stratified rankings are constructed for each instance, this results in $702 \times 2 = 1,402$ instance-level correlation samples, enabling a reliable empirical assessment of agreement with low randomness.

\paragraph{Metrics calculation.}
We set the hyper-parameter $\lambda_m$ in Eq.~\ref{eq:merge_score} as $0.2$, which considers the instance-related point-wise evaluation results provided by $S_{WPA}$. 
The weighted scoring points, point-wise alignment, penalty, and pseudo-human labels are generated by GPT-4o with temperature $0.5$.

\paragraph{Ablation Test.}
For ``Score Scale Reduction'' in the ablation test, we apply a simplified scale remapping to the evaluation scores. For Coarse 5-level and Checklist, the original ratings of $1$ and $2$ are collapsed into $1$, and ratings of $3$, $4$, and $5$ are mapped to $5$. For Coarse 3-level, the original ratings or matching scores of $0.5$ and $1$ are collapsed into $1$. 
For ``Importance Weight Disturbance'', we set the weights of all scoring points equally and randomly choose weights from $\{1,2,3\}$ to derive the results of $\rm WPA_{\rm avg}$, $\rm PCP_{\rm avg}$ and $\rm WPA_{\rm random}$, $\rm PCP_{\rm random}$, separately. 

\paragraph{Behavioral Analysis}
For three sub-figures in behavioral analysis, we adopt the following experimental settings.
In the first sub-figure, scores produced by different evaluation methods are linearly normalized to the $[0,1]$ range to account for differences in scoring scales.
In the second subplot, we assess ranking stability under noise by injecting zero-mean Gaussian noise with standard deviations $\sigma\in\{0.0,0.01,0.02,0.05,0.1,0.15,0.2\}$ into the evaluation scores. For each noise level, rankings are recomputed and compared against the original ranking using Kendall’s $\tau$.
In the third subplot, instances are partitioned into four length-based bins. For each non-empty bin, the score distribution is visualized using box plots, where the mean value is explicitly marked with a star symbol, and the spread of the distribution reflects the variance. 
\paragraph{Light-evaluators training.}
In the training of lightweight evaluators, we collect valid responses-scoring point pairs $(q,\hat{a},s_i)$ and the related alignment degree scores $y$ generated by GPT-4o. We select the English instances to construct an English-only dataset and split it into training, validation, and test sets, consisting of $\{17 407,2 536,2 331\}$ instances, respectively.

For training of BERT-based lightweight evaluators, we adopted BERT-base-uncased~\cite{devlin-etal-2019-bert} with batch size $64$ and learning rate $5e-5$. 
For training of decoder-only lightweight evaluators, we select Qwen2.5-0.5B-Instruct, Qwen3-1.7B, and Qwen3-4B~\cite{yang2025qwen3}. We adopt LoRA for parameter-efficient fine-tuning. The LoRA rank is set to $8$ with a scaling factor of $32$. LoRA adapters are applied to all attention projection matrices (\texttt{q\_proj}, \texttt{k\_proj}, \texttt{v\_proj}, and \texttt{o\_proj}). A dropout rate of $0.05$ is used in the LoRA layers, and no bias parameters are trained. The fine-tuning objective follows causal language modeling.
During training, the batch size is set to $2$, with gradient accumulation applied for $4$ steps to achieve a larger effective batch size. The model is optimized using a learning rate of $2e-5$ for $10$ training epochs.

\subsection{Gradient-free Prompt Optimization}
Few-shot chain-of-thought prompting often suffers from demonstration bias, where selected examples overfit to specific reasoning patterns and fail to generalize~\cite{gou2025mitigating}. To explore an optional mitigation strategy for our scoring-point generation (\S\ref{ScorePointGen}), we utilize a gradient-free prompt optimization method. We first identify several unexpected scoring points produced by the initial prompt $\mathcal{P}$, then manually revise them and attach brief notes explaining each modification. The original and revised points are provided to the LLM to induce an optimized prompt:
\begin{equation}
    \mathcal{P}_{\mathrm{optim}} = \mathrm{LLM}(\mathcal{P}, q, a, \{s_i, w_i\},\{s_i', w_i'\}),
\end{equation}
where $(q, a)$ is the associated question–answer pair. ${s_i, w_i}$ and ${s_i', w_i'}$ are the unexpected points and their corrected versions. Instead of directly inserting curated demonstrations, this trick lets the LLM abstract the correction patterns and incorporate them into $\mathcal{P}_{\mathrm{optim}}$, which could potentially alleviate the demonstration bias.
Since our primary contribution lies in the weighted importance multi-point evaluation framework, we do not emphasize this prompt optimization method in the main body.

\begin{figure*}[t]
\centering
\begin{minipage}{0.95\linewidth}
\lstset{style=aclprompt}
\begin{lstlisting}[caption={Prompt for Coarse 3-level}, label={lst:prompt_for_coarse_3-level}]
## Role:
Suppose you are a text quality evaluator.
## Objective:
- Given the [Question] and [Reference answer], evaluate the quality of a [Generated Answer] according to a reference answer. 
- Determine whether the [Generated Answer] covers the key information and output an alignment score of the [Generated Answer].
- Explain the reasons for the output alignment score.
## Skills:
- Content Matching: Identify if the [Generated Answer] directly or indirectly addresses the reference answer.
- Scoring: Assign an alignment score of 0, 0.5, or 1 based on how well the [Generated Answer] covers the reference answer.
## Workflow:
- Evaluate the coverage and matching degree of the [Generated Answer] for the reference answer and give an alignment score and the related reason.
- You may consider task-specific qualities (e.g., coherence, engagingness, groundedness, naturalness, and understandability in dialogue tasks) ONLY as auxiliary signals to help determine whether the Generated Answer semantically covers the reference answer.
- Please identify if the [Generated Answer] contains the reference answer explicitly or implicitly and give an alignment score as the following standard. 
The maximum alignment score is 1. The alignment score of the [Generated Answer] on the reference answer should be assigned 0, 0.5, or 1 points. 0 means that the [Generated Answer] does not include or omits the reference answer, 0.5 means that the [Generated Answer] partially includes the reference answer, and 1 means that the generated answer completely covers the reference answer. 
## Constraints:
- The alignment score can only be 0, 0.5, or 1. 
- If the generated answer only covers part of the reference answer, the alignment score for the reference answer should be 0.5. If it fully covers the reference answer, the alignment score should be 1. Otherwise, the alignment score should be 0.
- The alignment score should correspond to its specific explanation, avoiding situations where partial coverage is explained but the alignment score is given as 0 or 1, which represent no coverage or full coverage, respectively
- The contents of [Reference answer] must be used as the core basis.
- The [Generated Answer] may have a similar meaning to the reference answer but with different wording. As long as the [Generated Answer] conveys a similar meaning and contains no errors, it can be considered as effectively covering the reference answer.
## Output your response in the following JSON format. Do NOT output anything other than the JSON format:
    {
        "reason": "Explain which key information from the reference answer is covered, partially covered, or missing in the generated answer", 
        "rating": 0.5 
    }
\end{lstlisting}
\end{minipage}
\end{figure*}

\subsection{Prompts Details}\label{prompt_engeneering}
All steps in our framework and baselines are instantiated using LLMs~\cite{hurst2024gpt} to derive scoring points and conduct point-wise assessments under the evaluation protocol.
The prompt for scoring points generation is presented in Listing~\ref{lst:points_generation}, and the prompts for weighted point-wise alignment and point-wise conflict penalty are listed in Listing~\ref{lst:wpa} and Listing~\ref{lst:pcp}, respectively.
To ensure a fair comparison, all model outputs are constrained to JSON or list-based formats. Importantly, our primary contribution lies in the evaluation framework and metric design, rather than in prompt construction or optimization.

\begin{figure*}[t]
\centering
\begin{minipage}{0.95\linewidth}
\lstset{style=aclprompt}
\begin{lstlisting}[caption={Prompt for Multiple Scoring Points Generation and Weight Assignment}, label={lst:points_generation}]
    ## Role:
    You are a professional review analyst.
    ## Objective:
    - Extract critical scoring points from the given reference answer.
    - Each scoring point should correspond to one atomic contribution or claim that can be independently checked in a generated answer. 
    - Evaluate the importance of each scoring point in terms of its necessity for answering the question correctly, and assign an importance weight.
    ## Skills:
    - Have a deep understanding and analysis of the given input.
    - Be able to identify and extract key information points from a reference answer.
    - The extracted points could be a reference to evaluate the quality of a generated answer or response. 
    ## Workflow:
    1. Read and understand the given question and reference answer comprehensively.
    2. Extract key scoring points from the reference answer, ensuring that each point is a complete semantic unit and addresses different aspects of the reference answer.
    3. Reorganize the extracted scoring points, and merge points of the same aspect into one scoring point, noting that such points may be located in different contexts.
    4. Record each scoring point in the specified format and give it an integer weight from 1 to 3. The larger the weight, the more important the scoring point is.
    - Scoring Points with a weight of 3 are necessary conditions for a correct and complete answer to the question. They are highly relevant to the question and typically provide direct answers to the given question. These points are essential for answering the question, and without them, the answer may be incomplete or result in point deduction. 
    - Scoring Points with a weight of 2 are related to the question but are not particularly crucial. These points provide additional information or details that may enhance the answer's overall quality or depth. 
    - Scoring Points with a weight of 1 are not completely relevant to the question and will not have a significant impact even if they are deleted.
    5. Make sure not to include any extra content that does not conform to the specified output format.
    ## Constraints:
    - Scoring points must be provided strictly in the specified output format.
    - Each scoring point should correspond to one atomic contribution or claim that can be independently checked in a generated answer, such as motivation, background, method, experiment for a paper, and room, food, service, and  location for review.
    - Similar aspects of the reference answer should be summarized into a single scoring point.
    - Do not treat every sentence or clause as a separate scoring point, and avoid breaking up semantically coherent content into multiple scoring points.. Each scoring point should be an independent, complete, and meaningful evaluation unit.
    - The generated scoring points should cover the original reference answer and not omit any important information from the original reference answer. Content or illusions that are irrelevant to the reference answer must not be introduced.
    - The number of scoring points with each weight is not fixed and depends on the reference answer.
    - The weight of the points can only be 1, 2 or 3, which is assigned based on semantic importance.
    ## Output format:
    Point content must be enclosed in **double square brackets `[[...]]`**. Point weight** must be enclosed in double parentheses `((...))`.
    Output the scoring points strictly in the format below, with no extra text, explanation, or Markdown:
    - [[Text of first scoring point]] | ((3))
    - [[Text of second scoring point]] | ((2))
    - [[Text of third scoring point]] | ((1))
\end{lstlisting}
\end{minipage}
\end{figure*}


\begin{figure*}[t]
\centering
\begin{minipage}{0.95\linewidth}
\lstset{style=aclprompt}
\begin{lstlisting}[caption={Prompt for Weighted Point Alignment}, label={lst:wpa}]
    ## Role: You are a scoring-point coverage evaluator.

    ## Objective:
    - Evaluate the quality of the [Generated Answer] based on the provided [Question], [Reference Answer], and corresponding [Scoring Points].
    - Determine whether the [Generated Answer] covers the key information of each scoring point and calculate a matching score for each.
    - Explain the rationale behind the matching score for each scoring point.

    ## Skills:
    - **Understanding Scoring Points**: Able to comprehend the importance of each scoring point. Scoring points have weights of 1, 2, or 3, where 3 indicates key content, 2 indicates secondary content, and 1 indicates additional information.
    - **Content Matching**: Able to judge whether the [Generated Answer] directly or indirectly covers the scoring point.
    - **Scoring Ability**: Assign a matching score of 0, 0.5, or 1 based on the degree of coverage.

    ## Workflow:
    1. **Evaluate Each Scoring Point One by One**: Assess the [Generated Answer] against each scoring point in sequence.
    2. **Matching Score Allocation Criteria**:
    - If the [Generated Answer] does not cover or omits the scoring point, assign a score of 0.
    - If the [Generated Answer] partially covers the scoring point, assign a score of 0.5.
    - If the [Generated Answer] fully covers the scoring point, assign a score of 1.
    3. **Identify Match Type**:
    - Determine whether the [Generated Answer] includes the scoring point content explicitly or implicitly.
    - Consider semantic similarity: different wording can be accepted as valid coverage as long as the meaning is equivalent and accurate.
    4. **Justification**:
    - After assigning the matching score, provide a detailed explanation and error type for each scoring point.
    5. **Output Results**:
    - Each scoring point evaluation should include the point number, weight, matching score, and explanation.
    - Use a standardized format to output the results.

    ## Constraints:
    - Matching scores for each scoring point must be one of: 0, 0.5, or 1.
    - Matching scores must align with the justification. For example, a partially covered point cannot be given a 0 or 1.
    - The number of matching scores must equal the number of scoring points.
    - Semantically similar but differently worded content is allowed as valid coverage.
    - You must strictly follow the IDs and the order of the input scoring points. You must not create new scoring points, delete existing ones, or change the ids. Your output must contain exactly the same set of IDs as the input.

    ## Output Format:
    - Each scoring point evaluation must be **strictly valid JSON** with no extra text or markdown as follows:
    Example Format:
    {
        "point-wise scores": {
            "1": {
                "match_scores": 0.5 ,
                "explanation": "Justification for the assigned matching score",
                },
            ...
            "3": {
                "match_scores": 1 ,
                "explanation": "Justification for the assigned matching score",
                }
            }
        }
\end{lstlisting}
\end{minipage}
\label{prompt_for_WPA}
\end{figure*}

\begin{figure*}[t]
\centering
\begin{minipage}{0.95\linewidth}
\lstset{style=aclprompt}
\begin{lstlisting}[caption={Prompt for Point-wise Conflict Penalty}, label={lst:pcp}]
    ## Role:
    Suppose you are a text quality evaluator.
    ## Objective:
    - Given the [Question] [Reference answer] and the corresponding [Scoring points], evaluate the quality of a [Generated Answer] according to a standard answer containing several scoring points. 
    - Determine whether the [Generated Answer] contains anything that contradicts each scoring point and give a penalty score of 0 or 1.
    - Give the reasons for the penalty score assigned to each score point. 
    ## Skills:
    - Understanding Scoring Points: The end of each scoring point is marked with a weight of 1, 2 or 3 in ().
    - Scoring Points with a weight of 3 are critical elements in the reference answer. They are highly relevant to the question and typically provide direct answers to the given question. These points are essential for answering the question, and without them, the answer may be incomplete or result in point deduction.
    - Scoring Points with a weight of 2 are related to the question but are not particularly crucial. These points provide additional information or details that may enhance the answer's overall quality or depth. 
    - Scoring Points with a weight of 1 are not completely relevant to the question and will not have a significant impact even if they are deleted. 
    - Confliction identification: Identify if the [Generated Answer] directly or indirectly contradicts each scoring point.
    - Scoring: Assign a penalty score of 0 or 1 based on whether the [Generated Answer] contains conflict content with each scoring point.
    ## Workflow:
    - Identify the conflict between the [Generated Answer] and each scoring point in turn and give a penalty score of 0 or 1 and the related reason.
    - For each scoring point, please identify if the [Generated Answer] has anything that explicitly or implicitly conflicts with the score point and give a penalty score as the following standard.
    The penalty score for each scoring point could be assigned to 0 or 1. 
    0 means that the [Generated Answer] does not conflict with the score point.
    1 means that the [Generated Answer] partially includes content that contradicts the scoring point. 
    - Overall, each scoring point corresponds to a penalty score.
    ## Constraints:
    - For each scoring point, the penalty score can only be 0 or 1. 
    - For each scoring point, if the generated answer contains information that explicitly or implicitly conflicts with the score point, the penalty score should be 1. Otherwise, the penalty score should be 0.
    - The penalty score for each scoring point should correspond to its specific explanation, avoiding situations where the [Generated Answer] conflicts with the score point but the penalty score is given as 0, which represents no conflicts. 
    - The contents of [Reference answer] and [Scoring points] must be used as the core basis. A penalty score should be generated for each scoring point strictly. In other words, the number of penalty scores should equal the number of scoring points.
    - The number of penalty scores should be equal to the number of scoring points. 
    - Weight serves solely as descriptive metadata for each scoring point. It does NOT influence the determination of penalty scores.
    - Penalty scores must be assigned strictly based on whether the Generated Answer  contains explicit or implicit conflicts with the scoring point, regardless of weight.
    ## Output format:
    - Each scoring point evaluation must be **strictly valid JSON** with no extra text or markdown as follows:
    Example Format:
    {
        "point-wise penalty scores": {
            "1": {
                "penalty_scores": 0,
                "explanation": "Justification for the assigned penalty score",
                },
            ...
            }
     }
\end{lstlisting}
\end{minipage}
\end{figure*}

%% file: table/dataset_details.tex
\begin{table}[htbp]
\centering
\resizebox{\columnwidth}{!}{
\begin{tabular}{lcl}
\toprule
\textbf{Dataset} & \textbf{\# N} & \textbf{Category} \\
\midrule
StoryQA      & 14   & Narrative QA \\
ReviewSumm   & 134  & Opinion Summarization \\
MeetingSum   & 168  & Dialogue Summarization \\
FinancialQA  & 65   & Financial QA \\
PaperAssist  & 55   & Scientific Assistance \\
ConversMem   & 14   & Dialogue Memory \\
ContractQA   & 42   & Legal QA \\
LongStory    & 49   & Long-form Narrative \\
NewsSumm     & 11   & News Summarization \\
MultiDocQA   & 150  & Multi-Document QA \\
\bottomrule
\end{tabular}}
\caption{Statistics of evaluation datasets. \# N denotes the number of instances of each dataset. Categories reflect the primary domain or task focus of each dataset. }
\label{tab:dataset_stats}
\end{table}